\documentclass[12pt]{article}

\usepackage{natbib,hyperref}

\usepackage{amsmath}
\usepackage{amssymb}
\usepackage{graphicx}
\usepackage{float}

\usepackage{times}

\newenvironment{sciabstract}{%
\begin{quote} \bf}
{\end{quote}}

\newcounter{lastnote}

\title{Encoding Reality: Prediction-Assisted Cortical Learning Algorithm in Hierarchical Temporal Memory}

\author
{Fergal Byrne \\
\normalsize{HTM Theory Group, Dublin, Ireland}\\
\\
\normalsize{fergal@brenter.ie http://inbits.com}
}

\date{}

\begin{document} 


\baselineskip18pt


\maketitle


\begin{sciabstract}
In the decade since Jeff Hawkins proposed Hierarchical Temporal Memory (HTM) as a model of neocortical computation,
the theory and the algorithms have evolved dramatically. This paper presents a detailed 
description of HTM's Cortical Learning Algorithm (CLA), including for the first time a rigorous mathematical formulation of all aspects
of the computations. Prediction Assisted CLA (paCLA), a refinement of the CLA, is presented, which is both closer to the 
neuroscience and adds significantly to the computational power. Finally, we summarise the key functions of neocortex which 
are expressed in paCLA implementations. An Open Source project, Comportex, is the leading implementation of this evolving
theory of the brain.
\end{sciabstract}

\section{Introduction}
We present an up-to-date description of Hierarchical Temporal Memory (HTM) which includes a mathematical model of its 
computations, developed for this paper. The description and mathematics are presented here in order to provide an axiomatic
basis for understanding the computational power of each component in a HTM system, as well as a foundation for comparing
HTM computational models with empirical evidence of strcture and function in neocortex.

In particular, we demonstrate the following:

\begin{enumerate}
  \item A layer of HTM neurons automatically learns to efficiently represent sensory and sensorimotor inputs using semantic 
  encodings in the form of Sparse Distributed Representations (SDRs). These representations are robust to noise and missing or
  masked inputs, and generalise gracefully in a semantically useful manner.
  \item A HTM layer automatically learns high-dimensional transitions between SDRs, makes predictions of the future evolution
  of its inputs, detects anomalies in the dynamics, and learns high-order sequences of sensory or sensorimotor patterns.
  \item Transition Memory uses temporal and hierarchical context to assist recognition of feedforward patterns, enhancing 
  bottom-up input pattern recognition and providing for stabilisation in the face of uncertainty.
  \item HTM's Temporal Pooling models the learnable processing of fast-changing inputs in Layer 4 of cortex into slower-changing, stable
  representations in Layer 2/3 of sequences, orbits and trajectories of L4 SDRs.
\end{enumerate}

\section{Hierarchical Temporal Memory and the Cortical Learning Algorithm}

Hierarchical Temporal Memory was developed by Jeff Hawkins and Dileep George \citep{george2009} and substantially refined by Hawkins and his 
colleagues at 
Numenta. The most recent description produced by Numenta, \cite{HTMWhitePaper2011} has been updated by the author in \cite{HTMWhitePaper}. 
HTM is a model of cortex in which each region in a hierarchy 
performs an identical process on its own inputs, forming sequence memories of recognised spatial patterns. The Cortical Learning Algorithm 
(CLA) describes in detail how each region works.

\paragraph*{The HTM Model Neuron}
The model neuron in HTM is substantially more realistic and complex when compared to a point neuron in Artificial Neural Networks, such as the 
McCulloch-Pitts neuron \citep{mcculloch1943logical}, that led to Rosenblatt's perceptron. An ANN neuron simply
passes its inputs (weighted by synaptic strengths and summed) through a nonlinearity such as a sigmoid or rectifier. The extra complexity is 
intended to more carefully resemble the structure and function of real cortical neurons, while remaining simple compared with models based on
electrical characteristics.

\begin{figure}[H]
 \centering
 \includegraphics{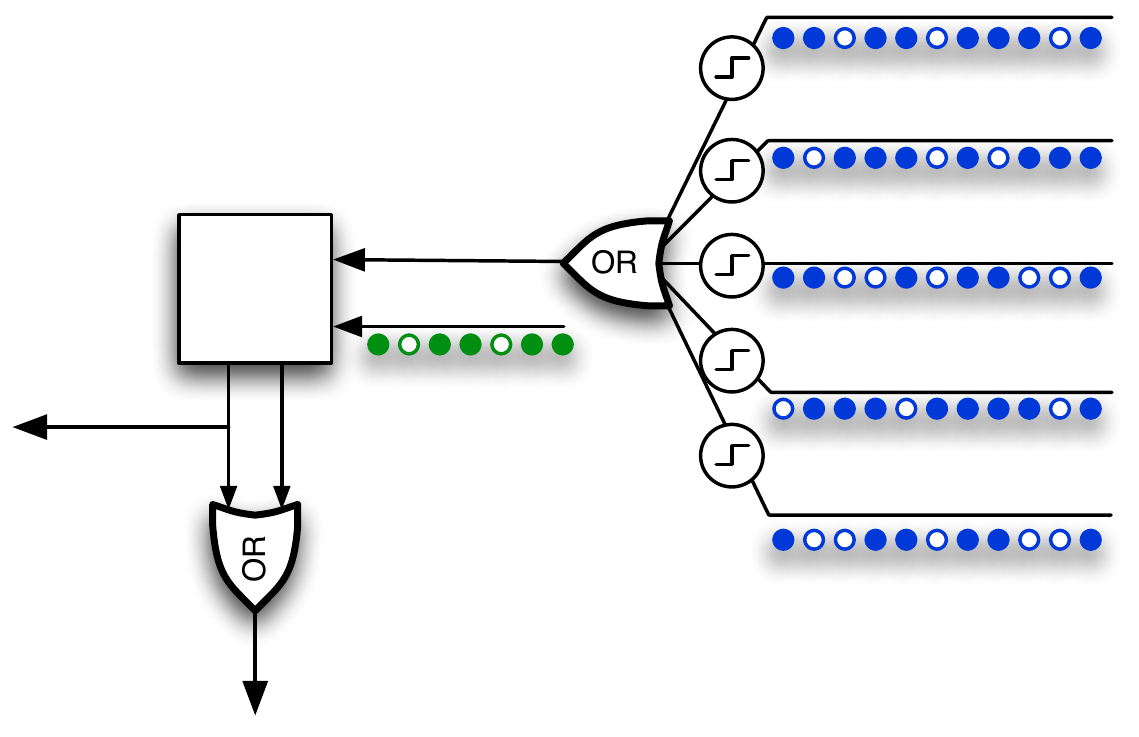}
 \caption{A HTM neuron. Feedforward inputs (green) appear on a proximal dendrite, where they enter the cell body or Soma. A set of distal dendrites 
 (blue) receive predictive contextual inputs from within the layer and from higher layers/regions. Each dendrite detects coincidental inputs and produces
 a dendritic spike only if its inputs exceed a threshold. From \cite{HTMWhitePaper}}
\end{figure}

The HTM model neuron has two kinds of dendrites, in order to process its two kinds of input. Feedforward inputs appear on a dendrite segment 
which is adjacent, or {\it proximal}, to the cell body ({\it soma}), and the sum of these inputs is directly fed into the soma. In addition, however,
the cell has a greater number of {\it distal} dendrite segments, each of which is an active unit capable of detecting the coincident activity of its own
inputs. The distal inputs are from nearby cells in the same layer, as well as cells in higher regions of the network. These cells provide
predictive context to assist the neuron in forming a decision to fire. Each distal segment learns to recognise its own set of contextual patterns, 
and provides input to the neuron only when sufficient input activity appears on its synapses.

For simplicity, CLA uses binary activation vectors to communicate between neurons. This is justified by empirical evidence
regarding the very high failure rate of individual synapses and the inherent noise found in biological neuron circuits.

\paragraph*{HTM Mini-columns} 
Real layers of cortex have been found to organise their neurons in mini-columns of about 30 cells, which
have strongly correlated feedforward response. CLA relies on an interpretation of this structure in order to model prediction and sequence
memory. In Hawkins' original design, the column and its contained cells play separate, but co-operative roles in the computational model. We
present here a more integrated design for the mini-column, which provides extra power in the computation and directly produces a semantic
interpretation of the representations. We'll return to mini-columns when describing prediction.

\begin{figure}[H]
 \label{fig:column}
 \centering
 \includegraphics{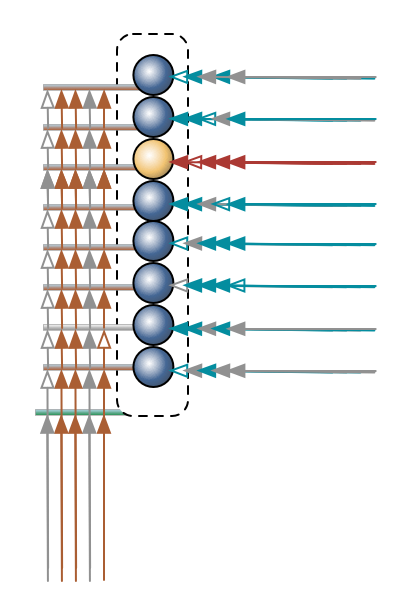}
 \caption{An HTM Column. The spheres are pyramidal model neurons, with their feedforward proximal dendrites at left and their individual 
 distal dendrites at right. The elongated vertical structure is an inhibitory column cell, which shares the same feedforward input as 
 its contained cells.}
\end{figure}

\paragraph*{Sparse Distributed Representations (SDRs)}
A final key feature of HTM is the Sparse Distributed Representation (SDR), which is a very sparse binary representation in which each active
bit has some semantic meaning. For further detail on SDRs, see \cite{SDRpaper}.

The next section describes the processes in CLA in more detail, and provides a full mathematical description in terms of vector operations.

\section{Pattern Memory (aka Spatial Pooling)}

We'll begin to describe the details and mathematics of HTM by describing the simplest operation in HTM's 
Cortical Learning Algorithm: Pattern Memory, also known as Spatial Pooling, forms a 
Sparse Distributed Representation from a binary feedforward input vector. Pattern Memory is a kind of learned 
spatial pattern recognition, capable of identifying and representing single inputs. 

We begin with a layer (a 1- or 2-dimensional array) of single neurons, which will form 
a pattern of activity aimed at efficiently representing the input vectors.

\begin{figure}[ht]
 \centering
 \includegraphics[scale=0.5,keepaspectratio=true]{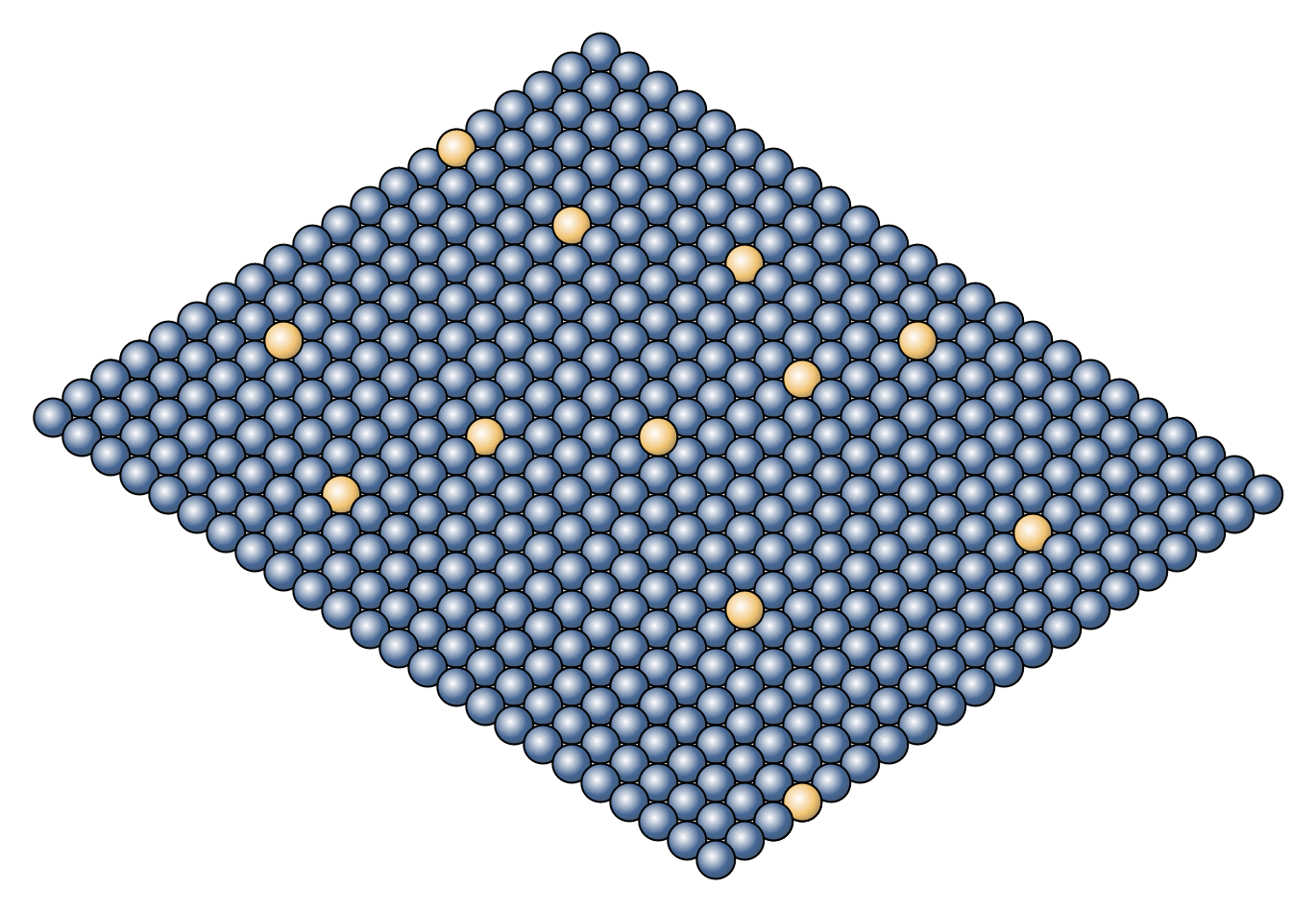}
 \caption{A single layer of CLA neurons, forming a Sparse Distributed Representation of active cells (yellow).}
\end{figure}

\subsection{Feedforward Processing on Proximal Dendrites}

The HTM model neuron has a single proximal dendrite, which is used to process and recognise 
feedforward or afferent inputs to the neuron. We model the entire feedforward input to a 
cortical layer as a bit vector ${\mathbf x}_{\textsc{ff}}\in\lbrace{0,1}\rbrace^{n_{\textsc{ff}}}$, 
where $n_{\textsc{ff}}$ is the width of the input.

The dendrite is composed of $n_s$ synapses which each act as a binary gate for a 
single bit in the input vector.  Each synapse $i$ has a {\it permanence} $p_i\in{[0,1]}$ which 
represents the size and efficiency of the dendritic spine and synaptic junction. The synapse 
will transmit a 1-bit (or on-bit) if the permanence exceeds a threshold $\theta_i$ (often a 
global constant $\theta_i = \theta = 0.2$). When this is true, we say the synapse is {\it connected}.

Each neuron samples $n_s$ bits from the $n_{\textsc{ff}}$ feedforward inputs, and so there 
are $\binom {n_{\textsc{ff}}} {n_{s}}$ possible choices of input for a single neuron. A single proximal 
dendrite represents a {\it projection} $\pi_j:\lbrace{0,1}\rbrace^{n_{\textsc{ff}}}\rightarrow\lbrace{0,1}\rbrace^{n_s}$, 
so a population of neurons corresponds to a set of subspaces of the sensory space. Each dendrite has an input 
vector ${\mathbf x}_j=\pi_j({\mathbf x}_{\textsc{ff}})$ which is the projection of the entire input into this 
neuron's subspace.

A synapse is connected if its permanence $p_i$ exceeds its threshold $\theta_i$. If we 
subtract ${\mathbf p}-{\vec\theta}$, take the elementwise sign of the result, and map 
to $\lbrace{0,1}\rbrace$, we derive the {\it binary connection vector }${\mathbf c}_j$ for the dendrite. Thus:

$$c_i=(1 + sgn(p_i-\theta_i))/2$$

The dot product $o_j({\mathbf x})={\mathbf c}_j\cdot{\mathbf x}_j$ now represents the {\it feedforward overlap} 
of the neuron with the input, ie the number of connected synapses which have an incoming activation potential. 
Later, we'll see how this number is used in the neuron's processing.

The elementwise product ${\mathbf o}_j={\mathbf c}_j\odot{\mathbf x}_j$ is the vector in the neuron's subspace 
which represents the input vector ${\mathbf x}_{\textsc{ff}}$ as "seen" by this neuron. This is known as 
the {\it overlap vector}. The length $o_j = \lVert{\mathbf o}_j\rVert_{\ell_1}$ of this vector corresponds to the 
extent to which the neuron recognises the input, and the direction (in the neuron's subspace) is that vector 
which has on-bits shared by both the connection vector and the input.

If we project this vector back into the input space, the result $\mathbf{\hat{x}}_j =\pi^{-1}({\mathbf o}_j)$ is this 
neuron's approximation of the part of the input vector which this neuron matches. If we add a set of such vectors, 
we will form an increasingly close approximation to the original input vector as we choose more and more neurons to 
collectively represent it.

\subsection{Sparse Distributed Representations (SDRs)}
We now show how a layer of neurons transforms an input vector into a sparse representation. From the above 
description, every neuron is producing an estimate $\mathbf{\hat{x}}_j $ of the input ${\mathbf x}_{\textsc{ff}}$, with 
length $o_j\ll n_{\textsc{ff}}$ reflecting how well the neuron represents or recognises the input. We form a sparse 
representation of the input by choosing a set $Y_{\textsc{sdr}}$ of the top $n_{\textsc{sdr}}=sN$ neurons, where $N$ is 
the number of neurons in the layer, and $s$ is the chosen {\it sparsity} we wish to impose (typically $s=0.02=2\%$).

The algorithm for choosing the top $n_{\textsc{sdr}}$ neurons may vary. In neocortex, this is achieved using a mechanism 
involving cascading {\it inhibition:} a cell firing quickly (because it depolarises quickly due to its input) activates 
nearby inhibitory cells, which shut down neighbouring excitatory cells, and also nearby inhibitory cells, which spread 
the inhibition outwards. This type of {\it local inhibition} can also be used in software simulations, but it is expensive 
and is only used where the design involves {\it spatial topology} (ie where the semantics of the data is to be reflected
in the position of the neurons). A more efficient {\it global inhibition} algorithm - simply choosing the top 
$n_{\textsc{sdr}}$ neurons by their depolarisation values - is often used in practise.

If we form a bit vector ${\mathbf y}_{\textsc{sdr}}\in\lbrace{0,1}\rbrace^N\textrm{ where 
} y_j = 1 \Leftrightarrow j \in Y_{\textsc{sdr}}$, we have a function which maps an input
${\mathbf x}_{\textsc{ff}}\in\lbrace{0,1}\rbrace^{n_{\textsc{ff}}}$ to a sparse 
output ${\mathbf y}_{\textsc{sdr}}\in\lbrace{0,1}\rbrace^N$, where the length of each output 
vector is $\lVert{\mathbf y}_{\textsc{sdr}}\rVert_{\ell_1}=sN \ll N$.

The reverse mapping or estimate $\mathbf{\hat{x}}$ of the input vector by the set $Y_{\textsc{sdr}}$ of neurons in the 
SDR is given by the sum:

$$\sum\limits_{j \in Y_{\textsc{sdr}}}{{\mathbf{\hat{x}}}_j} = 
\sum\limits_{Y_{\textsc{sdr}}}{\pi_j^{-1}({\mathbf o}_j)} = 
\sum\limits_{Y_{\textsc{sdr}}}{\pi_j^{-1}({\mathbf c}_j\odot{\mathbf x}_j)}= 
\sum\limits_{Y_{\textsc{sdr}}}{\pi_j^{-1}({\mathbf c}_j \odot \pi_j({\mathbf x}_{\textsc{ff}}))}= 
\sum\limits_{j \in Y_{\textsc{sdr}}}{\pi_j^{-1}({\mathbf c}_j) \odot {\mathbf x}_{\textsc{ff}}} $$

\begin{figure}[H]
 \centering
 \includegraphics[scale=0.8,keepaspectratio=true]{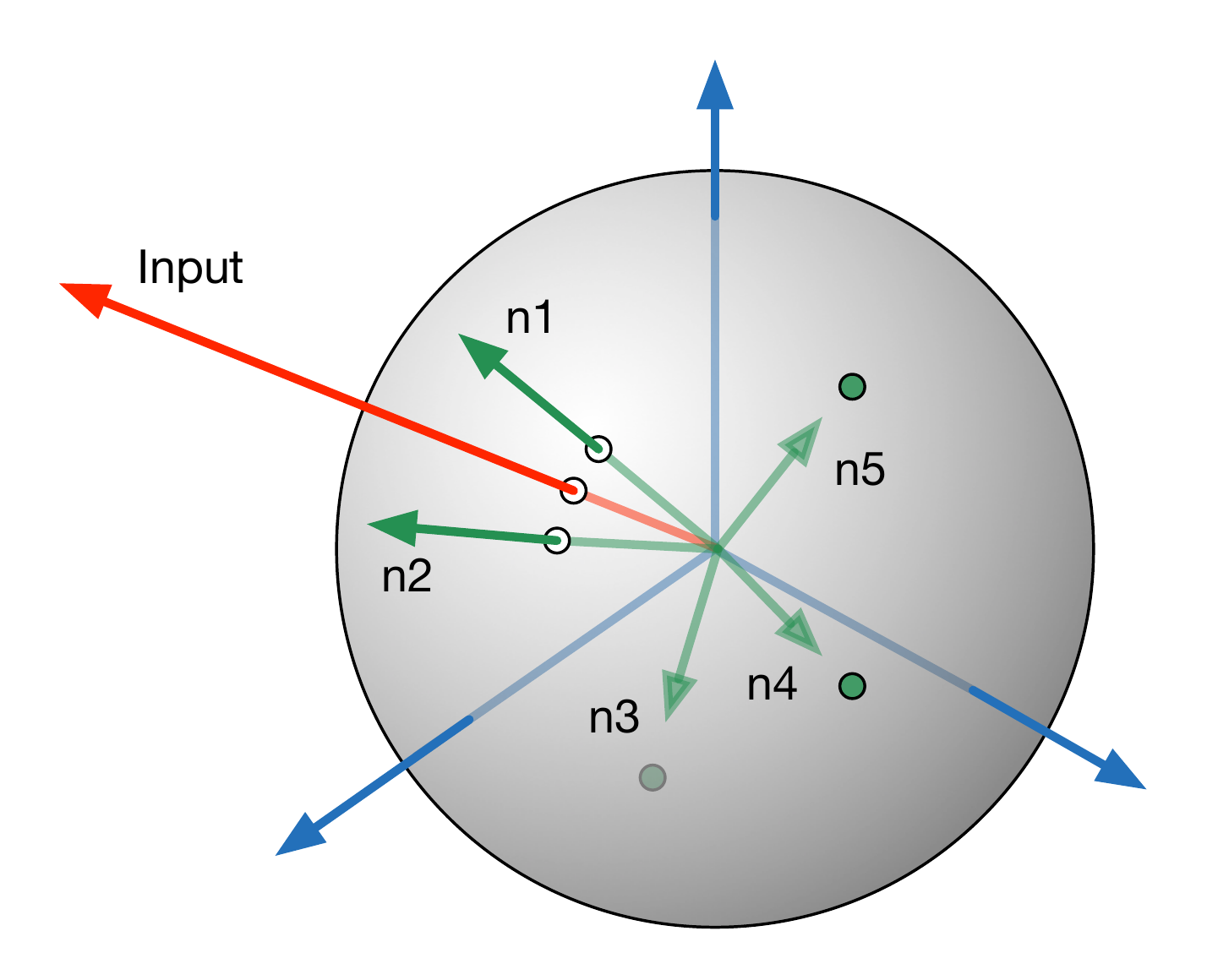}
 \caption{Visualisation of SDR Formation. The high-dimensional input space is projected to 3D for illustration. In red, the input
 vector has a particular orientation. The vectors n1-n5 represent a number of neurons in the layer, their length is proportional to
 the overlap with the input. Only neurons n1 and n2 have enough overlap to take part in the SDR, as these two are the best estimates
 of the input.}
 \label{fig:sphere-viz}
\end{figure}

\subsection{Matrix Form}
The above can be represented straightforwardly in matrix form. The projection 
$\pi_j:\lbrace{0,1}\rbrace^{n_{\textsc{ff}}} \rightarrow\lbrace{0,1}\rbrace^{n_s} $ can be represented as a matrix 
$\Pi_j \in \lbrace{0,1}\rbrace^{{n_s} \times\ n_{\textsc{ff}}} $.

Alternatively, we can stay in the input space ${\mathbb{B}}^{n_{\textrm{ff}}}$, and 
model $\pi_j$ as a vector $\vec\pi_j =\pi_j^{-1}(\mathbf 1_{n_s})$, ie 
where $\pi_{j,i} = 1 \Leftrightarrow (\pi_j^{-1}(\mathbf 1_{n_s}))_i = 1$.

The elementwise product $\vec{x_j} =\pi_j^{-1}(\mathbf x_{j}) = \vec{\pi_j} \odot {\mathbf x_{\textsc{ff}}}$ represents the 
neuron's view of the input vector $x_{\textsc{ff}}$.

We can similarly project the connection vector for the dendrite by elementwise multiplication: 
$\vec{c_j} =\pi_j^{-1}(\mathbf c_{j}) $, and thus $\vec{o_j}(\mathbf x_{\textsc{ff}}) = \vec{c_j} \odot \mathbf{x}_{\textsc{ff}}$ is the 
overlap vector projected back into $\mathbb{B}^{n_{\textsc{ff}}}$, and the dot product 
$o_j(\mathbf x_{\textsc{ff}}) = \vec{c_j} \cdot \mathbf{x}_{\textsc{ff}}$ gives the same {\it overlap} score for the neuron 
given $\mathbf x_{\textsc{ff}}$ as input. Note that $\vec{o_j}(\mathbf x_{\textsc{ff}}) =\mathbf{\hat{x}}_j $, the partial estimate of 
the input produced by neuron $j$.

We can reconstruct the estimate of the input by an SDR of neurons $Y_{\textsc{sdr}}$:

$$\mathbf{\hat{x}}_{\textsc{sdr}} = \sum\limits_{j \in Y_{\textsc{sdr}}}{{\mathbf{\hat{x}}}_j} = 
\sum\limits_{j \in Y_{\textsc{sdr}}}{\vec o}_j = 
\sum\limits_{j \in Y_{\textsc{sdr}}}{{\vec c}_j\odot{\mathbf x_{\textsc{ff}}}} = 
{\mathbf C}_{\textsc{sdr}}{\mathbf x_{\textsc{ff}}}$$

where ${\mathbf C}_{\textsc{sdr}}$ is a matrix formed from the ${\vec c}_j$ for $j \in Y_{\textsc{sdr}}$.

\subsection{Learning in HTM as an Optimisation Problem}
We can now measure the distance between the input vector $\mathbf x_{\textsc{ff}}$ and the reconstructed estimate 
$\mathbf{\hat{x}}_{\textsc{sdr}}$ by taking a norm of the diference. Using this, we can frame learning in HTM as an 
optimisation problem. We wish to minimise the estimation error over all inputs to the layer. Given a set of 
(usually random) projection vectors $\vec\pi_j$ for the N neurons, the parameters of the model are the 
permanence vectors $\vec{p}_j$, which we adjust using a simple Hebbian update model.

The update model for the permanence of a synapse $p_i$ on neuron $j$ is:

$$ p_i^{(t+1)} =
\begin{cases}
(1+\delta_ {inc})p_ i^ {(t)} & \text{if } j \in Y_ {\textrm{SDR}}\text{, }(\mathbf x_ j)_ i=1\text{ and } p_ i^ {(t)} \ge \theta_ i \\
(1-\delta_ {dec})p_ i^ {(t)} & \text{if } j \in Y_ {\textrm{SDR}} \text{ and (}(\mathbf x_ j)_ i=0 \text{ or }p_ i^ {(t)} < \theta_ i \text{)} \\
p_ i^ {(t)} & \text{otherwise} \\
\end{cases} $$

This update rule increases the permanence of {\it active synapses}, those that were connected to an active input when the cell became 
active, and decreases those which were either disconnected or received a zero when the cell fired. In addition to this rule, an external 
process gently {\it boosts} synapses on cells which either have a lower than target rate of activation, or a lower than target average 
overlap score.

In the visualisation above (see Figure \ref{fig:sphere-viz}), this will tend to move the vectors belonging to successful neurons closer to the input vector by increasing the number
of overlapping synapses, and it will also make the vectors more likely to remain active even in the face of noise in the input.

\subsection{Computational Power of SDR-forming Circuits}
An SDR is a form of k-winner-takes-all (k-WTA) representation. \cite{WTAMaass2000} proves that a single k-WTA gate has the same computational 
power as a polynomially larger multi-layer network of threshold artificial neurons, and that the soft (continuous) version can approximate 
any continuous function (exactly as a multilayer ANN network can - see \cite{maass1997networks}).

\section{Transition Memory - Making Predictions}
We saw how a layer of neurons learns to form a Sparse Distributed Representation (SDR) of an input pattern. In this section, 
we'll describe the process of learning temporal sequences.

We showed earlier that the HTM model neuron learns to recognise subpatterns of feedforward input on its proximal dendrites. This is 
somewhat similar to the manner by which a Restricted Boltzmann Machine can learn to represent its input in an unsupervised learning 
process. One distinguishing feature of HTM is that the evolution of the world over time is a critical aspect of what, and how, the 
system learns. The premise for this is that objects and processes in the world persist over time, and may only display a portion of 
their structure at any given moment. By learning to model this evolving revelation of structure, the neocortex can more efficiently 
recognise and remember objects and concepts in the world.

\subsection{Distal Dendrites and Prediction}

In addition to its one proximal dendrite, a HTM model neuron has a collection of {\it distal} (far) dendrite segments, or simply dendrites, 
which gather information from sources other than the feedforward inputs to the layer. In some layers of neocortex, these dendrites combine 
signals from neurons in the same layer as well as from other layers in the same region, and even receive indirect inputs from neurons in 
higher regions of cortex. We will describe the structure and function of each of these.

The simplest case involves distal dendrites which gather signals from neurons within the same layer.

Earlier, we showed that a layer of $N$ neurons converted an input vector $\mathbf x \in \mathbb{B}^ {n_ {\textrm{ff}}}$ into a 
SDR $\mathbf{y}_ {\textrm{SDR}} \in \mathbb{B}^ {N}$, with length $\lVert{\mathbf y}_ {\textrm{SDR}}\rVert_ {\ell_ 1}=sN \ll N$, 
where the sparsity $s$ is usually of the order of 2\% ($N$ is typically 2048, so the SDR $\mathbf{y}_ {\textrm{SDR}}$ will have 
40 active neurons).

The layer of HTM neurons can now be extended to treat its own activation pattern as a separate and complementary input for the next 
timestep. This is done using a collection of distal dendrite segments, which each receive as input the signals from other neurons 
in the layer itself. Unlike the proximal dendrite, which transmits signals directly to the neuron, each distal dendrite acts 
individually as an active coincidence detector, firing only when it receives enough signals to exceed its individual threshold.

We proceed with the analysis in a manner analogous to the earlier discussion. The input to the distal dendrite segment $k$ at time $t$ 
is a sample of the bit vector $\mathbf{y}_ {\textrm{SDR}}^ {(t-1)}$. We have $n_ {ds}$ distal synapses per segment, a permanence vector 
$\mathbf{p}_ k \in [0,1]^ {n_ {ds}}$ and a synapse threshold vector $\vec{\theta}_ k \in [0,1]^ {n_ {ds}}$, where 
typically $\theta_ i = \theta = 0.2$ for all synapses.

Following the process for proximal dendrites, we get the distal segment's connection vector $\mathbf{c}_ k$:

$$
c_ {k,i}=(1 + sgn(p_ {k,i}-\theta_ {k,i}))/2
$$

The input for segment $k$ is the vector $\mathbf{y}_ k^ {(t-1)} = \phi_ k(\mathbf{y}_ {\textrm{SDR}}^ {(t-1)})$ formed by the 
projection $\phi_ k:\lbrace{0,1}\rbrace^ {N-1}\rightarrow\lbrace{0,1}\rbrace^ {n_ {ds}}$ from the SDR to the subspace of the 
segment. There are $\binom {N-1} {n_ {ds}}$ such projections (there are no connections from a neuron to itself, so there 
are $N-1$ to choose from).

The overlap of the segment for a given $\mathbf{y}_ {\textrm{SDR}}^ {(t-1)}$ is the dot product 
$o_ k^ t = \mathbf{c}_ k\cdot\mathbf{y}_ k^ {(t-1)}$. If this overlap exceeds the threshold $\lambda_ k$ of the segment, 
the segment is active and sends a dendritic spike of size $s_ k$ to the neuron's cell body.

This process takes place before the processing of the feedforward input, which allows the layer to combine contextual knowledge of 
recent activity with recognition of the incoming feedforward signals. In order to facilitate this, we will change the algorithm for 
Pattern Memory as follows.

Each neuron $j$ begins a timestep $t$ by performing the above processing on its ${n_ {\textrm{dd}}}$ distal dendrites. This results
in some number $0\ldots{n_ {\textrm{dd}}}$ of segments becoming active and sending spikes to the neuron. The total predictive activation 
potential is given by:

$$
o_ {\textrm{pred},j}=\sum\limits_ {o_ k^ {t} \ge \lambda_ k}{s_ k}
$$

The predictive potential is combined with the overlap score from the feedforward overlap coming from the proximal dendrite to give the 
total activation potential:

$$
a_ j^ t=\alpha_ j o_ {\textrm{ff},j} + \beta_ j o_ {\textrm{pred},j}
$$

and these $a_ j$ potentials are used to choose the top neurons, forming the SDR $Y_ {\textrm{SDR}}$ at time $t$. The mixing factors 
$\alpha_ k$ and $\beta_ k$ are design parameters of the simulation.

\subsection{Learning Predictions}

We use a very similar learning rule for distal dendrite segments as we did for the feedforward inputs:

$$
{p_ {i,j}}^ {(t+1)} =
\begin{cases}
(1+\sigma_ {inc})p_ i^ {(t)} & \text {if cell } j \text{ active, segment } k \text{ active, synapse } i \text{ active} \\
(1-\sigma_ {dec})p_ i^ {(t)} & \text {if cell } j \text { active, segment } k \text{ active, synapse } i \text{ not active} \\
p_ i^ {(t)} & \text{otherwise} \\
\end{cases}
$$

Again, this reinforces synapses which contribute to activity of the cell, and decreases the contribution of synapses which don't. A 
boosting rule, similar to that for proximal synapses, allows poorly performing distal connections to improve until they are good enough 
to use the main rule.

\paragraph*{Interpretation}

We can now view the layer of neurons as forming a number of representations at each timestep. The field of predictive 
potentials $o_ {\textrm{pred},j}$ can be viewed as a map of the layer's confidence in its prediction of the next input. The field 
of feedforward potentials $o_ {\textrm{ff},j}$ can be viewed as a map of the layer's recognition of current reality. Combined, these 
maps allow for prediction-assisted recognition, which, in the presence of temporal correlations between sensory inputs, will improve the 
recognition and representation significantly.

We can quantify the properties of the predictions formed by such a layer in terms of the mutual information between the SDRs at 
time $t$ and $t+1$. .

A layer of neurons connected as described here is a Transition Memory, and is a kind of first-order memory of temporally 
correlated transitions between sensory patterns. This kind of memory may only learn one-step transitions, because the SDR is 
formed only by combining potentials one timestep in the past with current inputs.

Since the neocortex clearly learns to identify and model much longer sequences, we need to modify our layer significantly in order to 
construct a system which can learn high-order sequences. This is the subject of the next section.

\subsection{Higher-order Prediction}

The current Numenta Cortical Learning Algorithm (or CLA, the detailed computational model in HTM) separates feedforward and predictive stages 
of processing. A modification of this model (which we call {\it prediction-assisted recognition} or {\it paCLA}) combines these into a single 
step involving competition between highly predictive pyramidal cells and their surrounding columnar inhibitory sheaths.

Neural network models generally model a neuron as somehow "combining" a set of inputs to produce an output. This is based on the idea that 
input signals cause ion currents to flow into the neuron's cell body, which raises its voltage (depolarises), until it reaches a threshold level 
and fires (outputs a signal). paCLA also models this idea, with the added complication that there are two separate pathways (proximal and distal) 
for input signals to be converted into effects on the voltage of the cell. In addition, paCLA treats the effect of the inputs as a 
{\it rate of change} of potential, rather than as a final potential level as found in standard CLA.

\subsubsection*{Slow-motion Timeline of paCLA}

Consider a single column of pyramidal cells in a layer of cortex. Along with the set of pyramidal cells $\{P_1,P_2 .. P_n\}$, we also model 
each columnar sheath of inhibitory cells as a single cell $I$. All the $P_i$ and $I$ are provided with the same feedforward input vector 
$\mathbf{x}_t$, and they also have similar (but not necessarily identical) synaptic connection vectors $\mathbf{c}_{P_i}$ and $\mathbf{c}_{I}$ 
to those inputs (the bits of $\mathbf{x}_t$ are the incoming sensory activation potentials, while bit $j$ of a connection vector $\mathbf{c}$ 
is 1 if synapse $j$ is connected). The {\it feedforward overlap} $o^{\textrm{ff}}_{P_i}(\mathbf{x}_t) = \mathbf{x}_t \cdot \mathbf{c}_{P_i}$ 
is the output of the proximal dendrite of cell ${P_i}$ (and similarly for cell $I$).

In addition, each pyramidal cell (but not the inhibitory sheath) receives signals on its distal dendrites. Each dendrite segment acts separately 
on its own inputs $\mathbf{y}_k^{t-1}$, which come from other neurons in the same layer as well as other sublayers in the region (and from other 
regions in some cases). When a dendrite segment $k$ has a sufficient {\it distal overlap}, exceeding a threshold $\lambda_k$, the segment 
emits a dendritic spike of size $s_k$. The output of the distal dendrites is then given by:

$$o^{\textrm{pred}}=\sum\limits_{o_k^{t} \ge \lambda_k}{s_k}$$

The predictive potential is combined with the overlap score from the feedforward overlap coming from the proximal dendrite to give the 
{\it total depolarisation rate}:

$$d_j =  \frac{\partial V_j}{\partial t} = \alpha_j o^{\textrm{ff}}_{P_j} + \beta_j o^{\textrm{pred}}_{P_j}$$

where $\alpha_j$ and $\beta_j$ are parameters which transform the proximal and distal contributions into a rate of change of potential 
(and also control the relative effects of feedforward and predictive inputs). For the inhibitory sheath $I$, there is only the feedforward 
component $\alpha_I o^{\textrm{ff}}_I$, but we assume this is larger than any of the feedforward contributions $\alpha_j o^{\textrm{ff}}_{P_j}$ 
for the pyramidal cells.

Now, the time a neuron takes to reach firing threshold is inversely proportional to its depolarisation rate. This imposes an ordering of the 
set $\{P_1..P_n,I\}$ according to their (prospective) firing times $\tau_{P_j} = \gamma_P \frac{1}{d_j}$ (and $\tau_I = \gamma_I \frac{1}{d_I}$).

\subsubsection*{Formation of the SDR in Transition Memory}

Zooming out from the single column to a neighbourhood (or sublayer) $L_1$ of columns $C_m$, we see that there is a local sequence $\mathbb{S}$ 
in which all the pyramidal cells (and the inhibitory sheaths) would fire if inhibition didn't take place. The actual sequence of cells which do 
fire can now be established by taking into account the effects of inhibition.

Let's partition the sequence as follows:

$$\mathbb{S} = \mathbb{P}^{\textrm{pred}} \parallel \mathbb{I}^{\textrm{pred}} \parallel \mathbb{I}^{\textrm{ff}} 
\parallel \mathbb{P}^{\textrm{burst}} \parallel \mathbb{I}^{\textrm{spread}}$$

where:
\begin{enumerate}
\item $\mathbb{P}^{\textrm{pred}}$ is the (possibly empty) sequence of pyramidal cells in a highly predictive state, which fire before their 
inhibitory sheaths (ie $\mathbb{P}^ {\textrm{pred}} = \{P \mid \tau_P < \tau_{I_m}, P \in C_m\}$);
\item $\mathbb{I}^{\textrm{pred}}$ is the sequence of inhibitory sheaths which fire due to triggering by their contained predictively firing 
neurons in $\mathbb{P}^{\textrm{pred}}$ - these cells fire in advance of their feedforward times due to inputs from $\mathbb{P}^{\textrm{pred}}$;
\item $\mathbb{I}^{\textrm{ff}}$ is the sequence of inhibitory sheaths which fire as a result of feedforward input alone;
\item $\mathbb{P}^{\textrm{burst}}$ is the sequence of cells in columns where the inhibitory sheaths have just fired but their vertical 
inhibition has not had a chance to reach these cells (this is known as {\it bursting}) - ie 
$\mathbb{P}^{\textrm{burst}} =\{P~|~\tau_P < \tau_{I_m} + \Delta\tau_{\textrm{vert}}, P \in C_m\}$;
\item Finally, $\mathbb{I}^{\textrm{spread}}$ is the sequence of all the other inhibitory sheaths which are triggered by earlier-firing 
neighbours, which spreads a wave of inhibition imposing sparsity in the neighbourhood.
\end{enumerate}

Note that there may be some overlap in these sequences, depending on the exact sequence of firing and the distances between active columns.

\begin{figure}[H]
 \centering
 \includegraphics[scale=0.58,keepaspectratio=true]{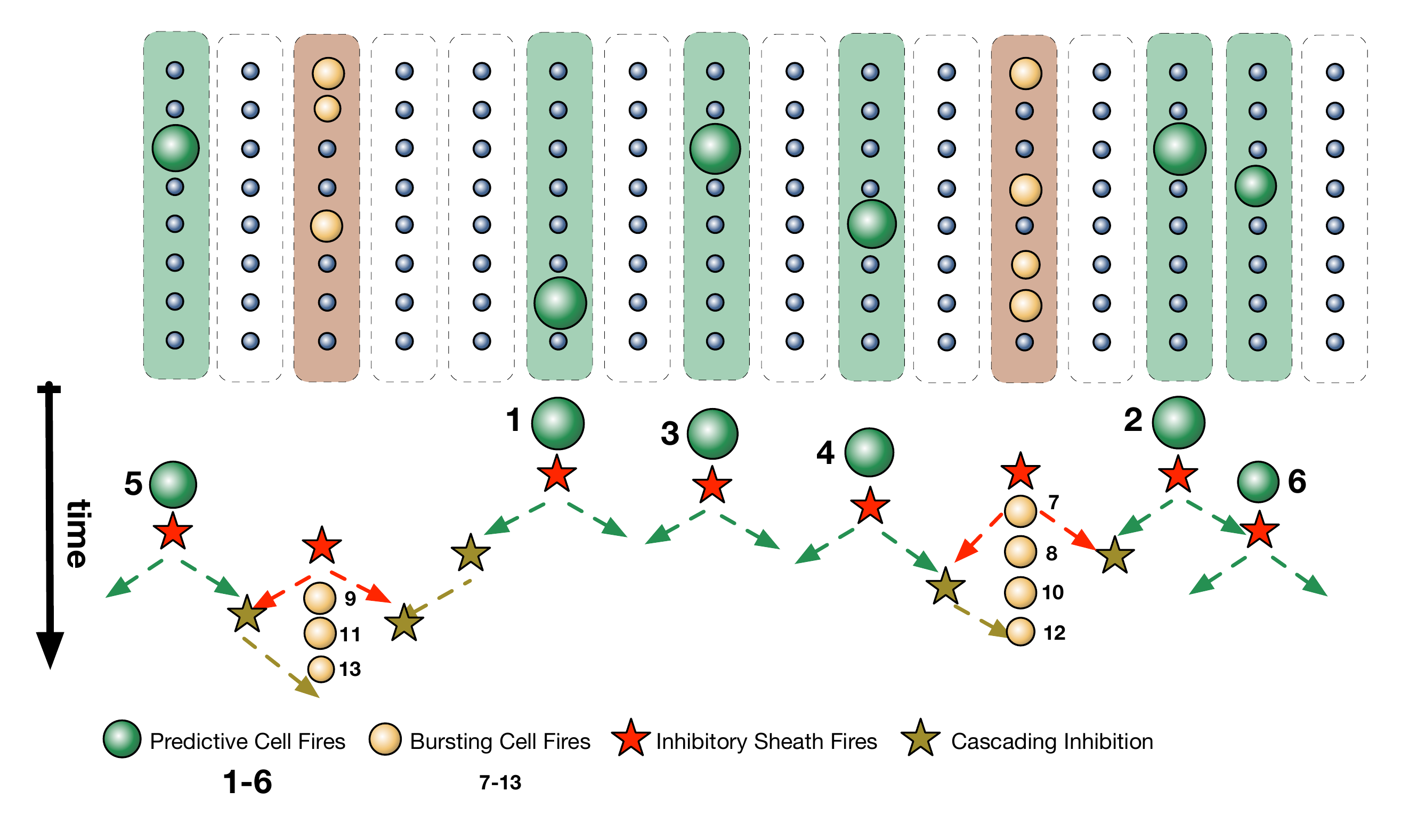}
 \caption{Schematic of SDR Formation by Inhibition. Top: A small slice of a CLA layer - columns activated by prediction assistance (green), 
 bursting due to feedforward input (orange), and inhibited (white). Bottom: Timeline (going downwards) showing activation and inhibition.}
\end{figure}

The output of a sublayer is the SDR composed of the pyramidal cells from $\mathbb{P}^{\textrm{pred}} \parallel \mathbb{P}^{\textrm{burst}}$ 
in that order. We say that the sublayer has {\it predicted perfectly} if $\mathbb{P}^{\textrm{burst}} = \emptyset$ and that the sublayer 
is {\it bursting} otherwise.

The cardinality of the SDR is minimal under perfect prediction, with some columns having a sequence of extra, bursting cells otherwise. 
The bursting columns represent feedforward inputs which were well recognised (causing their inhibitory sheaths to fire quickly) but less 
well predicted (no cell was predictive enough to beat the sheath), and the number of cells firing indicates the uncertainty of which 
prediction corresponds to reality. The actual cells which get to burst are representative of the most plausible contexts for the 
unexpected input.

\subsubsection*{Transmission and Reception of SDRs}

A sublayer $L_2$ which receives this $L_1$ SDR as input will first see the minimal SDR $\mathbb{P}^{\textrm{pred}}$ representing the 
perfect match of input and prediction, followed by the bursting SDR elements $\mathbb{P}^{\textrm{burst}}$ in decreasing order of 
prediction-reality match.

This favours cells in $L_2$ which have learned to respond to this SDR, and even more so for the subset which are also predictive due 
to their own contextual inputs (this biasing happens regardless of whether the receiving cells are proximally or distally enervated). The 
more sparse (well-predicted) the incoming SDR, the more sparse the activation of $L_2$.

When there is a bursting component in the SDR, this will tend to add significant (or overwhelming) extra signal to the minimal SDR, leading 
to high probability of a change in the SDR formed by $L_2$, because several cells in $L_2$ will have a stronger feedforward response to the 
extra inputs than those which respond to the small number of signals in the minimal SDR.

For example, in software we typically use layers containing 2,048 columns of 32 pyramidal neurons (64K cells), with a minimal column SDR of 
40 columns (c. 2\%). At perfect prediction, the SDR has 40 cells (0.06\%), while total bursting would create an SDR of 1280 cells. In between, 
the effect is quite uneven, since each bursting column produces several signals, while all non-bursting columns stay at one. Assuming some 
locality of the mapping between $L_1$ and $L_2$, this will have dramatic local effects where there is bursting.

The response in $L_2$ to bursting in its input will not only be a change in the columnar representation, but may also cause bursting 
in $L_2$ itself if the new state was not well predicted using $L_2$'s context. This will cause bursting to propagate downstream, from 
sublayer to sublayer (including cycles in feedback loops), until some sublayer can stop the cascade either by predicting its input or 
by causing a change in its external world which indirectly restores predictability.

Since we typically do not see reverberating, self-reinforcing cycles of bursting in neocortex, we must assume that the brain has learned 
to halt these cascades using some combination of eventual predictive resolution and remediating output from regions. Note that each 
sublayer has its own version of "output" in this sense - it's not just the obvious motor output of L5 which can "change the world". For 
example, L6 can output a new SDR which it transmits down to lower regions, changing the high-level context imposed on those regions 
and thus the environment in which they are trying (and failing somewhat) to predict their own inputs. L6 can also respond by altering its 
influence over thalamic connections, thus mediating or eliminating the source of disturbance. L2/3 and L5 both send SDRs up to higher 
regions, which may be able to better handle their deviations from predictability. And of course L5 can cause real changes in the world 
by acting on motor circuits.

\subsubsection*{How is Self-Stabilisation Learned?}

When time is slowed down to the extent we've seen in this discussion, it is relatively easy to see how neurons can learn to contribute 
to self-stabilisation of sparse activation patterns in cortex. Recall the general principle of Hebbian learning in synapses - the more 
often a synapse receives an input within a short time before its cell fires, the more it grows to respond to that input. 

Consider again the sequence of firing neurons in a sublayer:

$$\mathbb{S} = \mathbb{P}^{\textrm{pred}} \parallel \mathbb{I}^{\textrm{pred}} \parallel \mathbb{I}^{\textrm{ff}} 
\parallel \mathbb{P}^{\textrm{burst}} \parallel \mathbb{I}^{\textrm{spread}}$$

This sequence does not include the very many cells in a sublayer which do not fire at all, because they are contained either in columns 
which become active, but are not fast enough to burst, or more commonly they are in columns inhibited by a spreading wave from active 
columns. Let's call this set $\mathbb{P}^{\textrm{inactive}}$.

A particular neuron will, at any moment, be a member of one of these sets. How often the cell fires depends on the average amount of time 
it spends in each set, and how often a cell fires characteristically for each set. Clearly, the highly predictive cells in 
$\mathbb{P}^{\textrm{pred}}$ will have a higher typical firing frequency than those in $\mathbb{P}^{\textrm{burst}}$, while those 
in $\mathbb{P}^{\textrm{inactive}}$ have zero frequency when in that set. 

Note that the numbers used earlier (65536 cells, 40 cells active in perfect prediction, 1280 in total bursting) mean that the percentage 
of the time cells are firing on average is massively increased if they are in the predictive population. Bursting cells only fire once 
following a failure of prediction, with the most predictive of them effectively winning and firing if the same input persists.

Some cells will simply be lucky enough to find themselves in the most predictive set and will strengthen the synapses which will 
keep them there. Because of their much higher frequency of firing, these cells will be increasingly hard to dislodge and demote from 
the predictive state.

Some cells will spend much of their time only bursting. This unstable status will cause a bifurcation among this population. A portion 
of these cells will simply strengthen the right connections and join the ranks of the sparsely predictive cells (which will eliminate 
their column from bursting on the current inputs). Others will weaken the optimal connections in favour of some other combination of 
context and inputs (which will drop them from bursting to inactive on current inputs). The remainder, lacking the ability to improve 
to predictive and the attraction of an alternative set of inputs, will continue to form part of the short-lived bursting behaviour. 
In order to compete with inactive cells in the same column, these metastable cells will have to have an output which tends to feed 
back into the same state which led to them bursting in the first place. 

Cells which get to fire (either predictively or by bursting) have a further advantage - they can specialise their sensitivity to 
feedforward inputs given the contexts which caused them to fire, and this will give them an ever-improving chance of beating the 
inhibitory sheath (which has no context to help it learn). This is another mechanism which will allow cells to graduate from 
bursting to predictive on a given set of inputs (and context).

Since only active cells have any effect in neocortex, we see that there is an emergent drive towards stability and sparsity 
in a sublayer. Cells, given the opportunity, will graduate up the ladder from inactive to bursting to predictive when presented 
with the right inputs. Cells which fail to improve will be overtaken by their neighbours in the same column, and demoted back down 
towards inactive. A cell which has recently started to burst (having been inactive on the same inputs) will be reinforced in that 
status if its firing gives rise to a transient change in the world which causes its inputs to recur. With enough repetition, a cell 
will graduate to predictive on its favoured inputs, and will participate in a sparse, stable predictive pattern of activity in the
sublayer and its region. The effect of its output will correspondingly change from a transient restorative effect to a self-sustaining, 
self-reinforcing effect.

\subsection{Spatial/Columnar Interpretation of Transition Memory SDRs}

Since cells in each TM column share very similar feedforward response, we can just consider which columns contain active cells
when presented with each input. This columnar SDR will be very similar to the SDR formed by the Pattern Memory alone (ie without
prediction), differing only where the prediction has changed the outcome of the inhibition stage, favouring columns which have 
strongly predictive cells. The TM columnar SDR will be more invariant to occlusion or noise in the inputs, but will also potentially
hallucinate some inputs as it causes the layer to see what is expected rather than what is actually seen. It is likely that this balance
between error correction and hallucination is dynamically adjusted in real cortex.

\section{Sequence Memory - High-Order Sequences}

A CLA layer which has multi-cell columns is capable of learning high-order sequences of feedforward input patterns, ie sequences in which
the next input $ \mathbf {x}_ {t+1}$ can be predicted based on all the observed patterns $\{ \mathbf {x}_ {t-i} \vert 0 \leq i \leq k \}$ for some 
$k$ steps in the past, rather than just the current input $\mathbf {x}_t$. Thus, a layer which has seen the sequences $ABCD$ and $XBCY$ will
correctly predict $D$ after seeing $ABC$ and $Y$ after seeing $XBC$.

To explain this important function, consider the columns representing $B$ in the above sequences. In each column, one cell will have a distal
dendrite segment which receives inputs representing $A$, and another will have learned to recognise a previous $X$. So, while (essentially) the same 
columns become active for both $B$'s, the active {\it cells} in each case will be different. Thus the TM activation encodes an entire sequence
of patterns at every step. This chain of individual cell-level representations persists across multiple common inputs, such as $BC$ in this 
example, allowing the CLA to correctly predict the $D$ or $X$ as appropriate.

In addition, this allows the layer to distinguish between repeated patterns in a sequence, such as the $S$'s in $MISSISSIPPI$, notes in music,
or words in a sentence (eg 5 distinct repetitions of the word {\it in} have already appeared in this one). 

\section{Multiple levels of representation}

Note that a CLA layer is producing a number of representations of its inputs simultaneously, and these representations can be seen 
as nested one within another. 

\begin{figure}[H]
 \centering
 \includegraphics[scale=0.64,keepaspectratio=true]{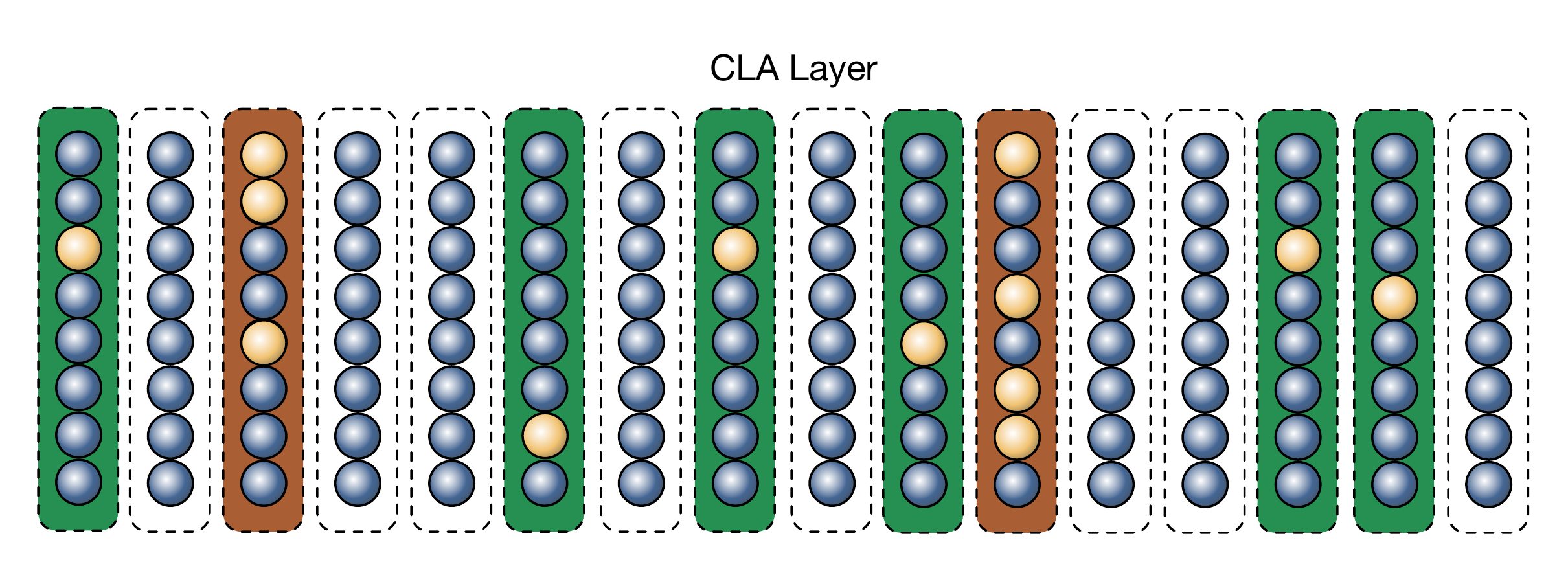}
 \caption{A small slice of a CLA layer. Predicted columns in green, bursting columns in orange. Active cells in yellow.}
\end{figure}

\paragraph*{Columnar SDR} The simplest and least detailed representation is the Columnar SDR, which is just a simple representation
of the pattern currently seen by the layer. This is what you would see if you looked down on the layer and just observed which columns
had active cells.

The number of patterns which can be represented is $\binom {N}{n_\mathrm{SDR}}$. In the typical software layer (2048 columns, 40 active),
we can have $\binom {2048} {40} = 2.37178*10^{84}$ SDRs. (See \citep{SDRpaper} for a detailed treatment of the combinatorics of SDRs).

\begin{figure}[H]
 \centering
 \includegraphics[scale=0.64,keepaspectratio=true]{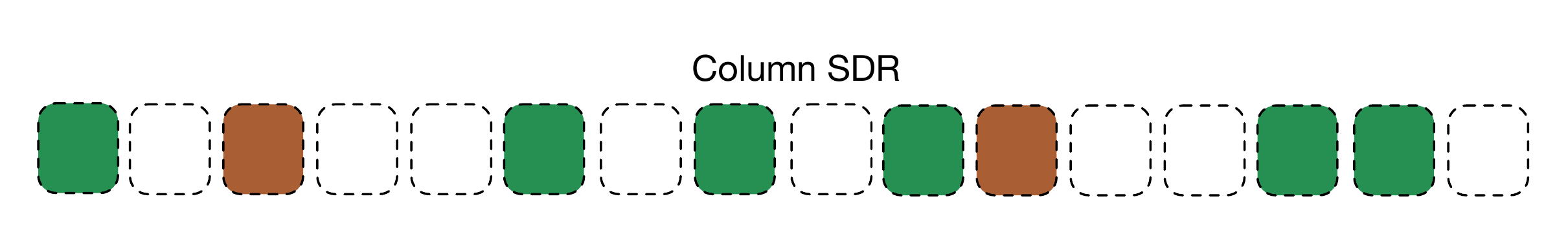}
 \caption{The Columnar SDR from the previous CLA layer.}
\end{figure}

\paragraph*{Cellular SDR} The cell-level SDR encodes both the Columnar SDR (if you ignore the choices of cells) and the context/sequence
in which it occurred. We can produce a one-cell-per-column SDR by choosing the most predictive cell in each active column (and choose randomly
in the case of bursting cells). In fact, this is how cells are chosen for learning in most implementations of CLA.

Interestingly, the capacity of this SDR is very large. For every Columnar SDR (ie for each spatial input), there are $n^{n_\mathrm{SDR}}$ 
distinct contexts, if each column contains $n$ cells. Again, in typical software, $n_\mathrm{SDR} = 40$, $n = 32$, so {\it each} feedforward input
can appear in up to $1.60694*10^{60}$ different contexts. Multiplying these, we get $3.8113*10^{144}$ distinct Cellular SDRs.

\paragraph*{Predicted/Bursting Columnar SDR} This more detailed SDR is composed of the sub-SDRs (or vectors) representing a) what was predicted
and confirmed by reality and b) what was present in the input but not well-predicted. The layer's combined output vector can thus be seen as
the sum of two vectors - one representing the correctly predicted reality and the other a perpendicular prediction error vector:

$$\mathbf{y}_{\mathrm {SDR}} = \mathbf{y}_{\mathrm {pred}} + \mathbf{y}_{\mathrm {burst}}$$

As we'll see in the next section, this decomposition is crucial to the process of Temporal Pooling, in which a downstream layer can 
learn to stably represent a single representation by learning to recognise successive $\mathbf{y}_{\mathrm {pred}}$ vectors.

\begin{figure}[H]
 \centering
 \includegraphics[scale=0.64,keepaspectratio=true]{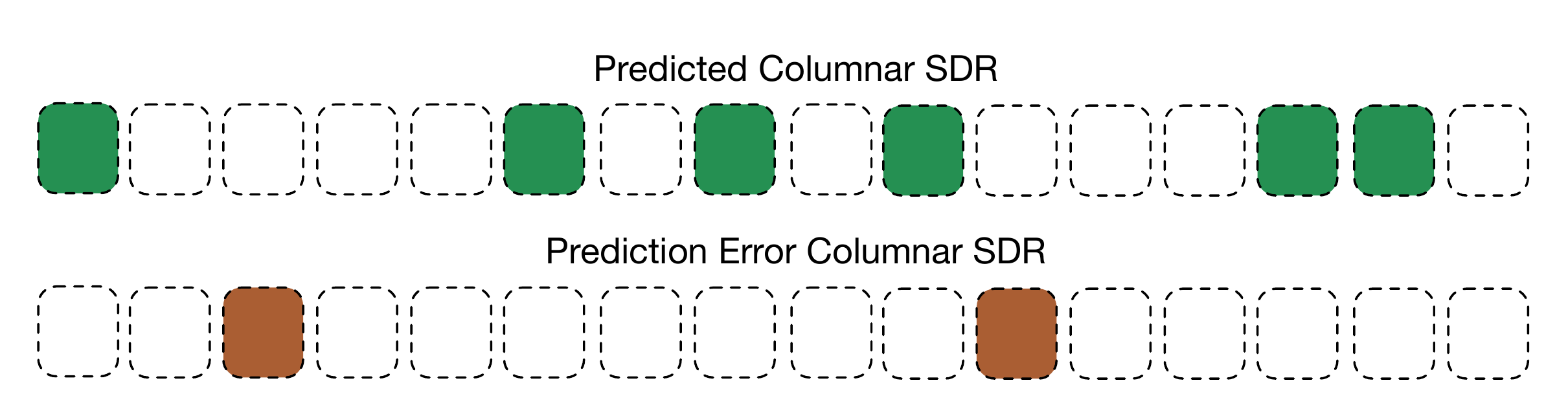}
 \caption{Columnar SDRs split into predicted (top) and prediction error (bottom) columnar SDRs}
\end{figure}

\paragraph*{Predicted/Bursting Cellular SDR} This is the cellular equivalent of the previous SDR (equivalently the previous SDR is the
column-level version of this one). This SDR encodes the precise sequence/context identity as well as the split between predicted and
prediction error vectors. In addition, looked at columnwise, the error SDR is actually the union of all the vectors representing how 
the input and prediction differed, thus forming a cloud in the output space whose volume reflects the confusion of the layer.

As noted earlier, the size, or ${\ell_ 1}$ norm, of the Predicted/Bursting Cellular SDR varies dramatically with the relative number 
of predicted vs bursting columns. In a typical CLA software layer, $40 \leq \lVert{\mathbf y}_ {\textrm{SDR}}\rVert_ {\ell_ 1} \leq 1280$, 
a 32x range.

\begin{figure}[H]
 \centering
 \includegraphics[scale=0.64,keepaspectratio=true]{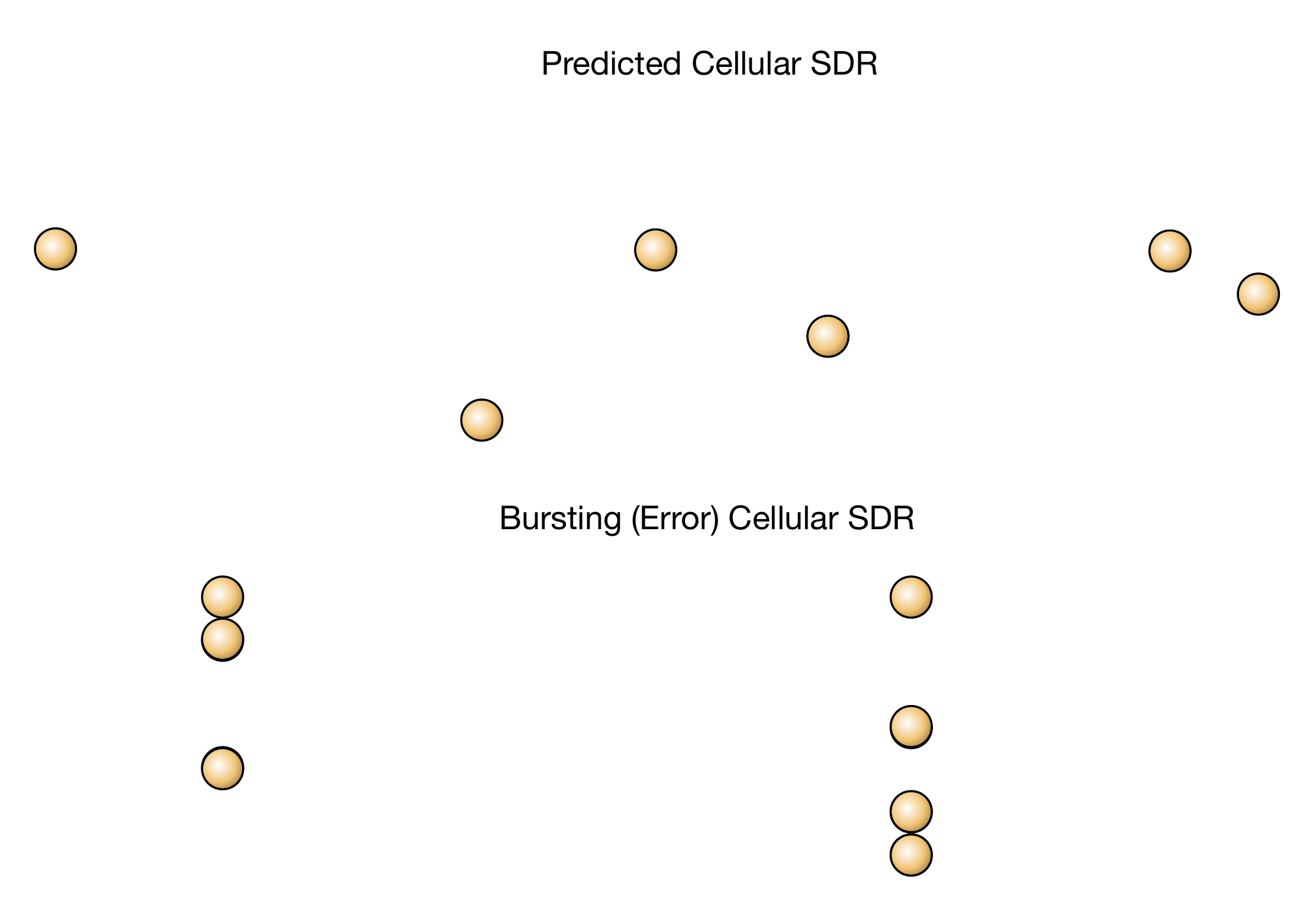}
 \caption{Cellular SDRs for predicted input (top) and prediction error (bottom)}
\end{figure}

\paragraph*{Prediction-ordered SDR Sequences} Even more detail is produced by treating the SDR as a {\it sequence} of individual activations,
as we did earlier when deriving the sequence:

$$\mathbb{S} = \mathbb{P}^{\textrm{pred}} \parallel \mathbb{P}^{\textrm{burst}}$$

Each of the two subsequences is ordered by the activation levels of the individual cells, in decreasing order. Each thus represents a
sequence of recognitions, with the most confident recognitions appearing earliest.

\begin{figure}[H]
 \centering
 \includegraphics[scale=0.64,keepaspectratio=true]{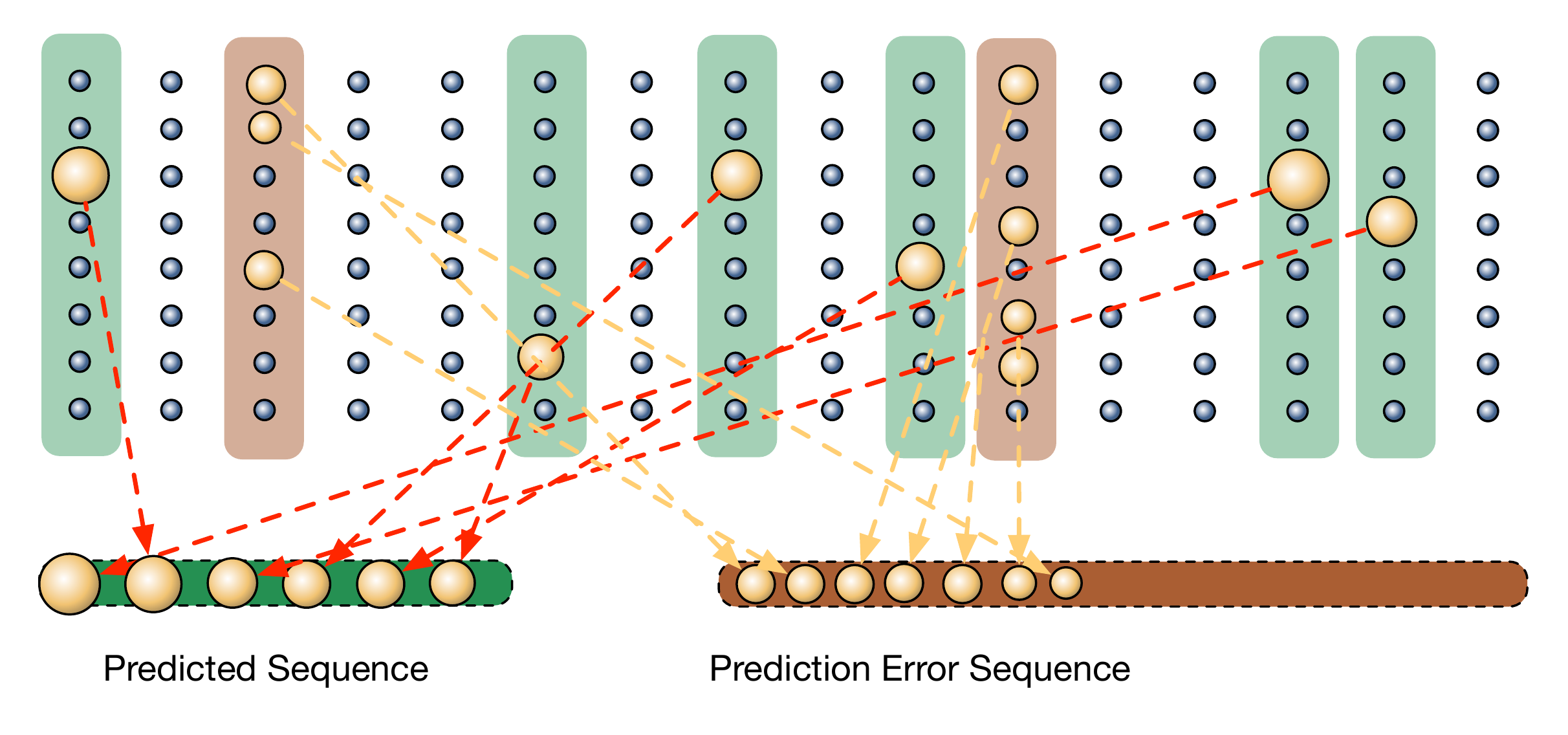}
 \caption{The SDR represented as a sequence for the predicted input (green) and the prediction error (orange).}
\end{figure}

\section{Temporal Pooling: from single- to multi-layer models}

One well-understood aspect of the structure of the neocortex is the hierarchical organisation of visual cortex. The key
feature of this hierarchy is that the spatial and temporal scale of receptive fields increases from low-level to high-level
regions. 

\begin{figure}[H]
 \centering
 \includegraphics[scale=0.7,keepaspectratio=true]{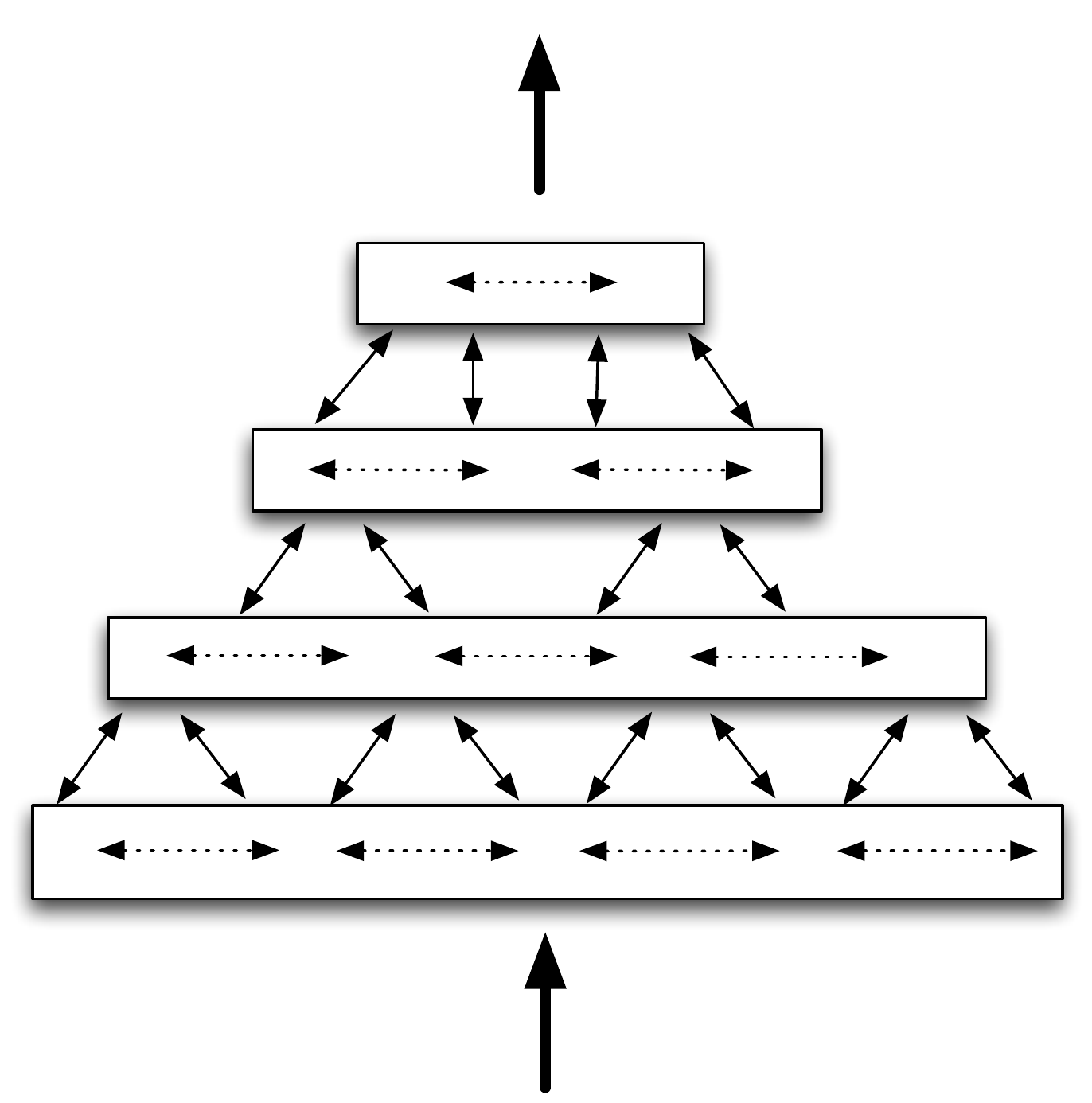}
 \caption{Schematic of cortical hierarchy. From \citep{HTMWhitePaper}}
\end{figure}

Early versions of HTM resembled Artificial Neural Networks, or Deep Learning Networks, in having a single layer for each region
in the hierarchy \cite{george2009}. The current CLA as described in \citep{HTMWhitePaper} continues this design, modelling only
a single layer, equivalent to Layer 2/3 in cortex in each region. The latest developments in HTM involve a new mechanism called
Temporal Pooling, which models both Layer 4 and Layer 2/3. This section describes Temporal Pooling and its role in extracting 
hierarchical spatiotemporal information from sensory and sensorimotor inputs.

\begin{figure}[H]
 \centering
 \includegraphics[scale=0.63,keepaspectratio=true]{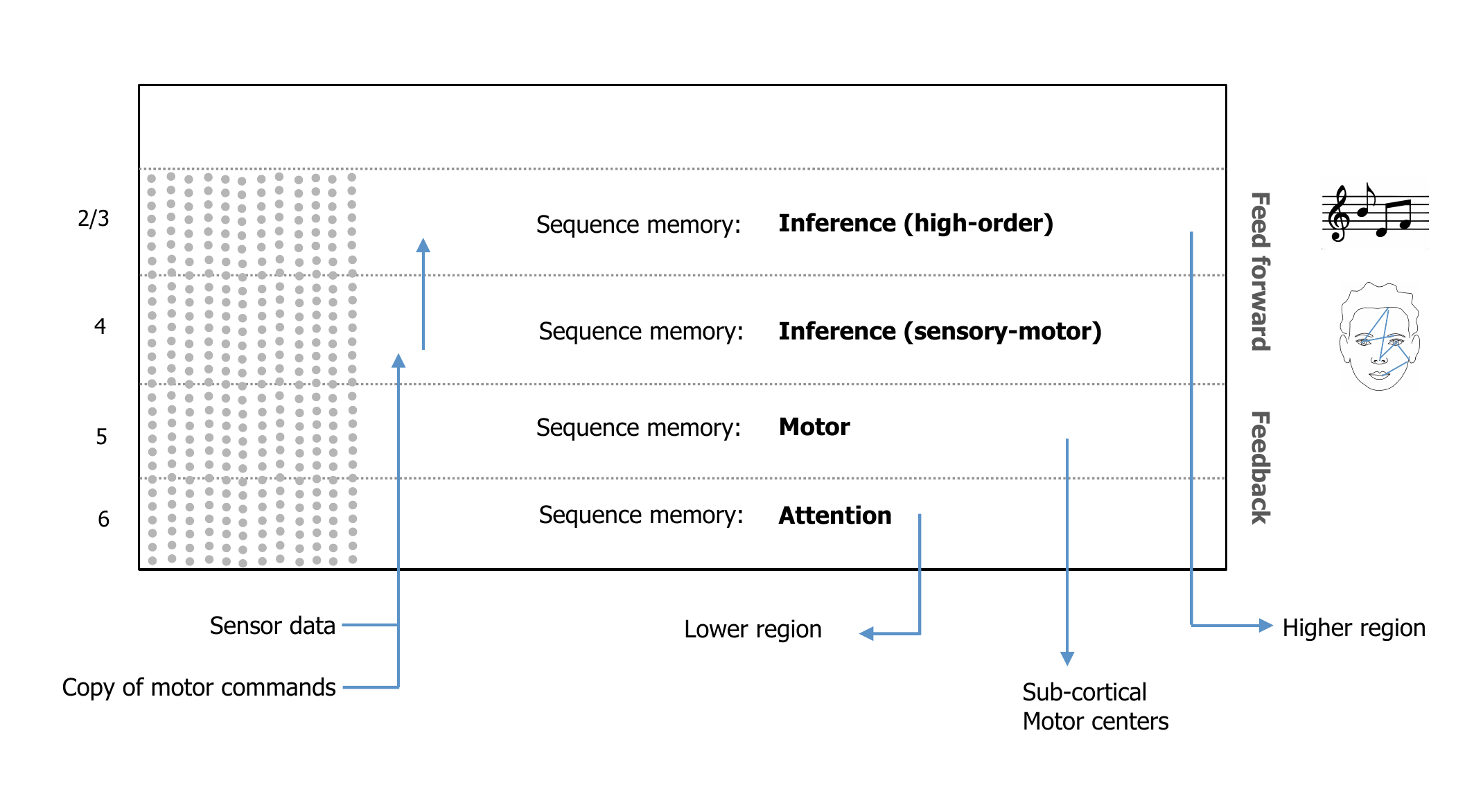}
 \caption{Hawkins' Multilayer Schematic}
\end{figure}

Hawkins proposes that each layer is performing a similar task of learning sequences of its inputs, but with differences in
the processing in each layer. For this discussion, the layers of interest are Layer 4, which receives direct sensorimotor feedforward
input, and Layer 2/3, which receives as input the output of Layer 4, producing the representation which is passed up the hierarchy.

The idea of Temporal Pooling is as follows. Layer 4 is receiving a stream of fast-changing sensorimotor inputs, and uses its Transition
Memory to form predictions of the next input in the stream. If this succeeds, the sequence of SDRs produced in Layer 4 will each be a sparse 
set of predictive cells. Temporal Pooling cells in Layer 2/3 which have proximal synapses to many L4 cells in a particular sequence will
then repeatedly become active as L4 evolves through the sequence, and will form a stable, slowly-changing representation of the sequence
undergone by L4.

\begin{figure}[H]
 \centering
 \includegraphics[scale=0.71,keepaspectratio=true]{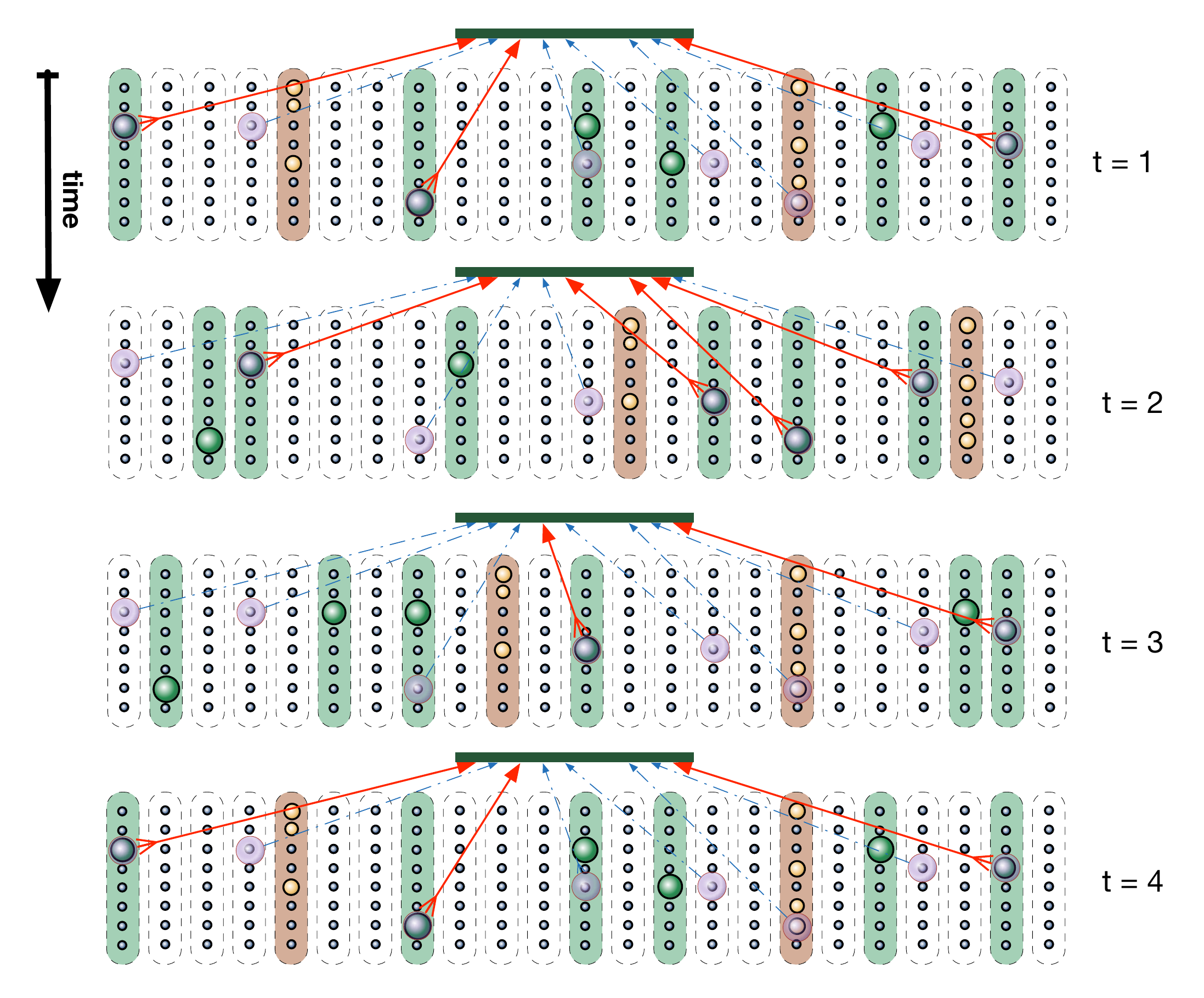}
 \caption{Sketch of Temporal Pooling. Four successive snapshots are shown. Above each CLA Layer 4 slice is a proximal dendrite belonging to
 a Layer 2/3 cell which has learnt to temporally pool over this sequence in L4. The pooling cell is connected to 8 L4 cells (purple), which
 together provide between 2 and 4 active inputs (red arrows) each timestep in the sequence. The L2/3 cell will thus stay active while this 
 sequence is predictable in L4.}
\end{figure}

\paragraph*{Learning in Temporal Pooling}
A simple extension of the Pattern Memory learning rule is sufficient to explain Temporal Pooling learning. Recall the original rule, 
the update model for the permanence of a synapse $p_i$ on neuron $j$ is:

$$ p_i^{(t+1)} =
\begin{cases}
(1+\delta_ {inc})p_ i^ {(t)} & \text{if } j \in Y_ {\textrm{SDR}}\text{, }(\mathbf x_ j)_ i=1\text{ and } p_ i^ {(t)} \ge \theta_ i \\
(1-\delta_ {dec})p_ i^ {(t)} & \text{if } j \in Y_ {\textrm{SDR}} \text{ and (}(\mathbf x_ j)_ i=0 \text{ or }p_ i^ {(t)} < \theta_ i \text{)} \\
p_ i^ {(t)} & \text{otherwise} \\
\end{cases} $$

Temporal Pooling simply uses different values for $\delta_ {inc}$ and $\delta_ {dec}$ depending on whther the input is from a predictive 
or bursting neuron in L4. For predicted neurons, $\delta_ {inc}$ is increased, and $\delta_ {dec}$ is decreased, and for bursting neurons
$\delta_ {inc}$ is decreased and $\delta_ {dec}$ is increased. This causes the pooling neuron to preferentially learn sequences of predicted
SDRs in L4.

\subsection*{Extension to Cycles and Trajectories in L4}
Hawkins' original description of Temporal Pooling \cite{TPNewIdeas} referred to {\it sequences} in L4, but since then it appears that
his conception of L4 sequence memory involves low- or zero-order memory rather than the long high-order sequences learned in the CLA we've
already been describing. 

It is likely that real cortex exploits the spectrum of L4 sequence capacity in order to match the dynamics of each region's sensory inputs. 

\section{Summary and Resources}

This theory aims to combine a reasonably simple abstraction of neocortical function with several key computational features which
we believe are central to understanding both mammalian and artificial intelligence. Using a simple but powerful mathematical description
of the paCLA algorithms, we can reason about their computational power and learning capabilities. This paper also provides a sound
basis for extending the theory in new directions. Indeed, we are developing a new, multilayer model of neocortex based on the current work.

Comportex \citep{Comportex} is an Open Source implementation of HTM/CLA which demonstrates most of the theory presented here, including both
paCLA and Temporal Pooling. For other resources on HTM, we recommend the website of the open source community at \cite{NumentaOrg}.

\bibliography{htm}
\bibliographystyle{plainnat}

\end{document}